\pdfoutput=1

\documentclass[11pt]{article}

\usepackage[preprint]{acl}
\usepackage{times}
\usepackage{latexsym}
\usepackage[T1]{fontenc}
\usepackage[utf8]{inputenc}
\usepackage{microtype}
\usepackage{inconsolata}
\usepackage{graphicx,xcolor}
\usepackage{subcaption}
\usepackage{multirow}
\usepackage{booktabs}
\usepackage{lscape}
\usepackage{pxrubrica}
\usepackage{url}

\title{Joint Extraction Matters: Prompt-Based Visual Question Answering for Multi-Field Document Information Extraction}

\author{Mengsay Loem \and Taiju Hosaka \\
  Sansan, Inc. \\
  \texttt{\{mengsay.loem, hosaka\}@sansan.com}
  }

\begin{document}
\maketitle

\begin{abstract} 
Visual question answering (VQA) has emerged as a flexible approach for extracting specific pieces of information from document images. However, existing work typically queries each field in isolation, overlooking potential dependencies across multiple items. This paper investigates the merits of extracting multiple fields jointly versus separately. Through experiments on multiple large vision language models and real-world datasets, we show that jointly extracting fields often improves accuracy—especially when the fields share strong numeric or contextual dependencies. We further analyze how performance scales with the number of requested items and use a regression-based metric to quantify inter-field relationships. Our results suggest that multi-field prompts can mitigate confusion arising from similar surface forms and related numeric values, providing practical methods for designing robust VQA systems in document information extraction tasks.
\end{abstract}

\section{Introduction}
Prompt-based visual question answering (VQA) methods~\cite{radford2021learning, Qwen2VL, grattafiori2024llama3herdmodels, antol2015vqa, Tanaka_Nishida_Yoshida_2021, Appalaraju_Tang_Dong_Sankaran_Zhou_Manmatha_2024} have recently emerged as a flexible and efficient approach for document information extraction~\cite{Shao_PromptingLLMsVQA, Yihao_VDoc2022}. 
Unlike earlier methods that rely on architectures explicitly designed for layout analysis or bounding-box detection, such as LayoutLM~\cite{xu2020layoutlmv2, huang2022layoutlmv3} and DETR~\cite{carion2020detr}, prompt-based VQA leverages large vision language models (VLMs) that respond to natural language prompts. 
This enables direct extraction of any desired information from document images (e.g., vendor name, total amount, transaction date from a receipt image). 
With this paradigm, a single model can be adapted to multiple tasks or fields simply by modifying the prompt, reducing the need for extensive model retraining or re-engineering.

\begin{figure}[t]
    \centering
    \includegraphics[width=0.48\textwidth]{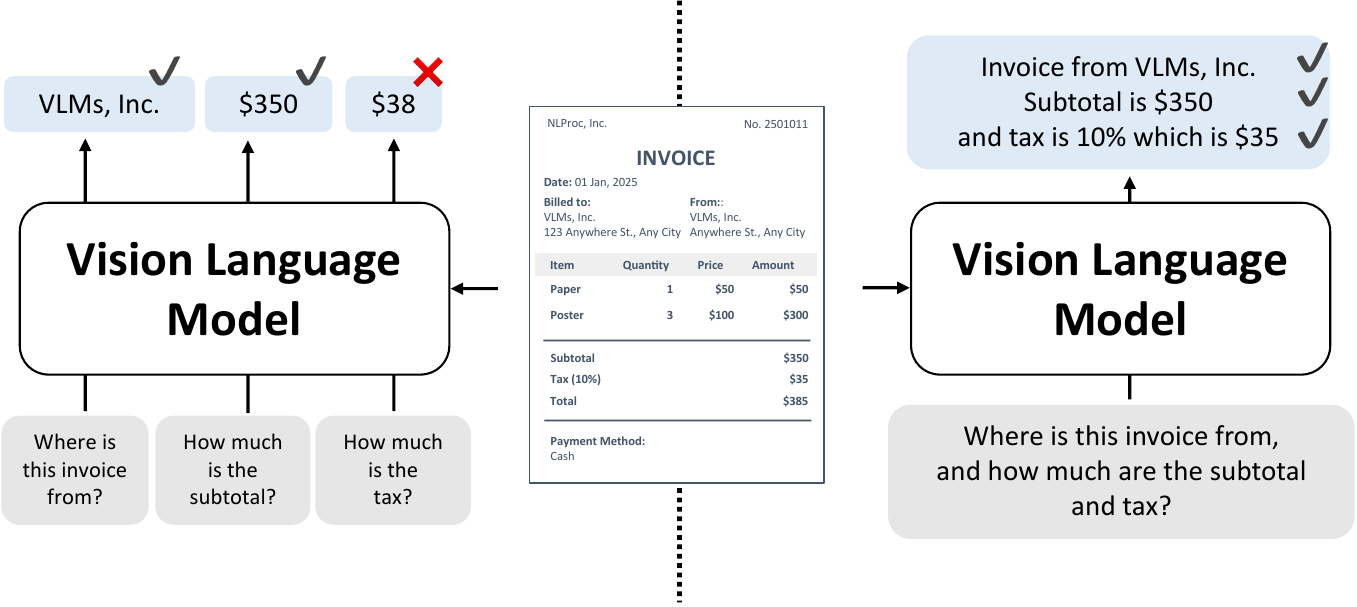}
    \caption{
    Overview of VQA-based information extraction framework using vision language models. 
    \textit{Left}: Each field is queried separately (\textit{separate extraction}). 
    \textit{Right}: A single prompt requests multiple fields simultaneously (\textit{joint extraction}).
    }
    \label{fig:overview}
\end{figure}

A key factor influencing the performance of prompt-based VQA methods is \textbf{how the prompt is formulated}. 
Recent studies have shown that the structure and wording of prompts can significantly impact model accuracy and consistency~\cite{jiang2020lm, pmlr-v202-zhang23m, lester-etal-2021-power, wu-etal-2022-idpg}. 
For example, structured prompts with explicit guidance often improve model outputs compared to open-ended queries~\cite{yang2023dynamicpromptingunifiedframework, loem-etal-2023-exploring, he2024doespromptformattingimpact}. 
This suggests that the way we formulate prompts—whether querying for fields separately or jointly—can directly affect extraction performance (Figure~\ref{fig:overview}).

Real-world documents (e.g., receipts, invoices, and forms) typically contain multiple, interrelated fields that need to be extracted for downstream applications~\cite{tang-etal-2024-multiple}. 
This complexity raises an important question for both industry practitioners and researchers:
\begin{quote}
    \emph{Is it better to query each field separately (\textit{separate extraction}), or can a single prompt requesting multiple fields (\textit{joint extraction}) yield superior accuracy?}
\end{quote}
On one hand, separate queries provide a straightforward pipeline, isolating the extraction of each field. 
However, our preliminary analyses suggest that visually or textually similar fields (e.g., multiple numeric entries) are prone to confusion when extracted separately, as the model may struggle to distinguish between them. 
On the other hand, a single prompt requesting \emph{multiple} fields simultaneously could leverage inter-field relationships (e.g., the correlation between \emph{tax}, \emph{subtotal}, and \emph{total amount}) allowing the model to infer more accurate and context-aware results.

To address this trade-off, we conduct a systematic investigation of \textbf{separate} versus \textbf{joint} extraction within a prompt-based VQA framework. 
Specifically, we evaluate three large vision language models (Qwen2-VL-2B Instruct, Llama-3.2-11B Vision Instruct, and GPT-4o) across multiple real-world document datasets. 
We measure both document- and field-level accuracy and introduce a regression-based metric to quantify the extent to which different fields depend on one another.

Our central hypothesis is that \textit{joint extraction} will be most beneficial when fields share similar surface forms or exhibit strong relational dependencies (e.g., \emph{subtotal} and \emph{tax} in numeric receipts). 
Conversely, if certain fields (e.g., \emph{store name} and \emph{date}) are relatively independent, a single multi-field prompt may provide only marginal benefits. 
Our experiments confirm this hypothesis: \textit{joint extraction} substantially improves performance for numerically or contextually intertwined fields, while the advantage remains modest for fields that are inherently distinct.

In summary, our contributions are as follows:
\begin{itemize}
    \item We systematically compare \textbf{separate} versus \textbf{joint} prompt-based VQA for multi-field document information extraction across multiple VLMs and datasets.
    \item We introduce a quantified \textbf{field-dependence measure} to analyze when joint extraction delivers the most substantial improvements.
    \item We provide \textbf{practical insights} into the role of prompt formulation in document VQA, demonstrating how grouping related or similar fields in a single prompt can enhance extraction accuracy and efficiency.
\end{itemize}

\section{Related Work}
Recent advances in document image understanding have shown that jointly modeling textual content, spatial layout, and visual cues significantly boosts performance across tasks such as information extraction and table structure analysis.
For instance, the LayoutLM family of models~\cite{Xu_2020LayoutLM, xu2020layoutlmv2, huang2022layoutlmv3} and LaTr~\cite{biten2022latr} integrate these modalities within Transformer-based architectures, achieving state-of-the-art accuracy on various benchmarks. 
These approaches can better capture the structure of documents by encoding both textual tokens and their 2D positions on the page, leading to improved entity recognition and form parsing.

Another major development has been the shift toward prompt-based VQA as a flexible strategy for document information extraction~\cite{Shao_PromptingLLMsVQA, Yihao_VDoc2022}, leveraging large VLMs pre-trained on massive image-text corpora~\cite{pmlr-v139-radford21a, bai2023qwenvlversatilevisionlanguagemodel, Qwen2VL, grattafiori2024llama3herdmodels}. 
Unlike layout-centric methods, these VLMs can directly respond to natural language prompts such as “What is the total amount on this receipt?” without extensive architecture-specific fine-tuning. 
This flexibility allows zero- or few-shot adaptation to various document types and information fields.

However, most existing work on the prompt-based VQA focuses on extracting one field per query, overlooking potential dependencies among multiple interrelated fields. 
While recent studies have proposed architecture refinements to deal with multiple questions in a single prompt~\cite{tang-etal-2024-multiple}, the exact benefits of querying multiple fields jointly, compared to separately, remain understudied. 
In this work, we aim to fill that gap by systematically evaluating separate versus joint extraction under various conditions. 
We seek to clarify when and why multi-field prompts confer advantages and provide insights beyond the specifics of any single architecture or training paradigm.

\begin{table}[t]
\centering
\resizebox{0.48\textwidth}{!}{%
\begin{tabular}{ll}
\hline
\textbf{Dataset} & \textbf{Focused Fields} \\ \hline
CORDv2           & Subtotal, Tax Amount, Service Charge, \\
                 & Total Price, Cash Payment Amount, \\
                 & Change Amount, Credit Card Payment Amount, \\
                 & Count of Menu Items, Quantity \\
\hline
SROIE Task 3     & Company Name, Address, \\
                 & Transaction Date, Total Amount \\
\hline
FUNSD            & Sender (From), Reciever (To), Title, Date \\
                 & Address, Telephone, Telefax \\
\hline
\end{tabular}
}
\caption{Key information fields from each dataset.}
\label{tab:dataset_items}
\end{table}

\begin{table*}[t]
\centering
\small
\begin{tabular}{lcccccc}
\toprule
\multirow{2}{*}{\textbf{Dataset}} 
& \multicolumn{2}{c}{\textbf{Qwen2-VL-2B}} 
& \multicolumn{2}{c}{\textbf{Llama-3.2-11B}} 
& \multicolumn{2}{c}{\textbf{GPT-4o}} \\
& \textbf{Sep.} & \textbf{Joint} 
& \textbf{Sep.} & \textbf{Joint} 
& \textbf{Sep.} & \textbf{Joint} \\
\midrule
CORDv2 & 0.77 & \textbf{0.83} & 0.40 & \textbf{0.71} & 0.96 & \textbf{0.98} \\
SROIE  & 0.94 & \textbf{0.95} & 0.81 & \textbf{0.83} & 0.97  & \textbf{0.98} \\
FUNSD  & 0.57  & \textbf{0.71}            & 0.73  & \textbf{0.79}          & 0.79 & \textbf{0.97} \\
\bottomrule
\end{tabular}
\caption{Document-level accuracy across three datasets. 
\textbf{Sep.} denotes separate extraction; \textbf{Joint} denotes joint extraction.}
\label{tab:doc_level_accuracy}
\end{table*}

\section{Experiments}
\subsection{Models and Datasets}
To systematically compare \textbf{separate} versus \textbf{joint} extraction in prompt-based VQA, we conduct experiments using three large vision language models across three real-world document datasets. 
Our goal is to evaluate how different prompting strategies affect extraction accuracy across varying model sizes and document types.

We use the following three models, each capable of handling image inputs and responding to natural language prompts with text outputs:
\begin{itemize}
    \item Qwen2-VL-2B-Instruct \footnote{\url{https://huggingface.co/Qwen/Qwen2-VL-2B-Instruct}}
    \item Llama-3.2-11B-Vision-Instruct \footnote{\url{https://huggingface.co/meta-llama/Llama-3.2-11B-Vision-Instruct}}
    \item GPT-4o \footnote{\url{https://platform.openai.com/docs/models/gpt-4o}}
\end{itemize}

We evaluate performance on three real-world document datasets, each representing different structured document types. Table~\ref{tab:dataset_items} summarizes the key information fields in each dataset.

\begin{itemize}
    \item \textbf{CORDv2}~\cite{park2019cord}: A dataset of scanned receipts with multi-level semantic labels. It contains 800 training images and 100 images each for development and testing. The dataset includes fields such as \textit{Subtotal}, \textit{Tax}, \textit{Service}, \textit{Total}, \textit{CreditCard}, \textit{Cash}, and \textit{Change}. Our experiments focus on numeric and payment-related fields, and all reported results use the test set.
    \item \textbf{SROIE Task 3}~\cite{Huang_2019SROIE}: A dataset from the SROIE 2019 competition that involves extracting key financial information from scanned receipts, including \textit{Company Name}, \textit{Address}, \textit{Transaction Date}, and \textit{Total Amount}. We use the official test set of 347 images for evaluation.
    \item \textbf{FUNSD-VQA}: A VQA-adapted variant\footnote{\url{https://huggingface.co/datasets/munish0838/funsd-vqa}} of the FUNSD dataset~\cite{jaume2019FUNSD} designed for document-based question answering. It consists of 149 preprocessed question-answer pairs for form-based document analysis, covering structured fields such as text boxes, form labels, and key-value pairs.
\end{itemize}

\subsection{Prompting Strategies}
\label{subsec:prompt_strategies}

To purely evaluate the capacity of models and compare \textit{separate} versus \textit{joint} extraction, we conduct our main experiments under a zero-shot setting. 
This approach minimizes confounding effects from example selection 
or incremental learning. 
For completeness, we provide 4-shot results in the Appendix~\ref{appendix:few-shot} as additional reference.

\paragraph{Separate Extraction.}
We issue an individual query to the VLM for each target field in a document. 
For example, to extract \textit{Subtotal}, \textit{Tax}, and \textit{Total}, 
we send three prompts:
\begin{quote}
\small
\texttt{Prompt 1: "Given the following image of a receipt, extract the Subtotal."}\\
\texttt{Prompt 2: "Given the following image of a receipt, extract the Tax."}\\
\texttt{Prompt 3: "Given the following image of a receipt, extract the Total."}
\end{quote}

\begin{table*}[t]
\centering
\small
\begin{tabular}{lp{12cm}}
\hline
\textbf{Field} & \textbf{Model's Output} \\ \hline
Tax &The receipt shows a total amount of 51,300 and a tax price of 10\%. We can calculate the tax price as follows:
Tax Price = Total Amount x Tax Rate
= 51,300 x 0.10
= 5,130 \\ \hline
Change & The receipt shows a total amount of 80,500 and a cash payment of 100,000. To calculate the change, we subtract the total from the payment amount: 100,000 - 80,500 = 19,500. This calculation reveals that the change amount is 19,500. \\ \hline
\end{tabular}
\caption{Example outputs (partial) from Llama-3.2-11B-Vison-Instruct with joint extraction.}
\label{tab:Joint_examples}
\end{table*}

\paragraph{Joint Extraction.}
We also evaluate extracting multiple fields in a single query. 
For instance:
\begin{quote}
\small
\texttt{"Given the following image of a receipt, extract the Subtotal, Tax, and Total."}
\end{quote}

\begin{table}[t!]
    \centering
    \resizebox{0.48\textwidth}{!}{%
    \begin{tabular}{lcccccc}
        \toprule
        \textbf{Item}
        & \multicolumn{3}{c}{\textbf{Llama-3.2-11B-Vision-Instruct}}
        & \multicolumn{3}{c}{\textbf{Qwen2-VL-2B-Instruct}} \\
        \cmidrule(lr){2-4} \cmidrule(lr){5-7}
        & \textbf{Separate} & \textbf{Joint} & \textbf{$\Delta$}
        & \textbf{Separate} & \textbf{Joint} & \textbf{$\Delta$} \\
        \midrule
        Subtotal    & 0.69 & 0.72 & +0.03 & 0.92 & 0.93 & +0.01 \\
        Total       & 0.74 & 0.75 & +0.01 & 0.96 & 0.98 & +0.02 \\
        Tax         & 0.65 & 0.79 & +0.14 & 0.79 & 0.89 & +0.11 \\
        CreditCard  & 0.93 & 0.93 & 0.00  & 0.87 & 0.88 & +0.01 \\
        Quantity    & 0.60 & 0.81 & +0.21 & 0.76 & 0.96 & +0.20 \\
        Cash        & 0.67 & 0.93 & +0.26 & 0.87 & 0.92 & +0.05 \\
        Change      & 0.89 & 0.94 & +0.05 & 0.92 & 0.98 & +0.06 \\
        Service     & 0.83 & 0.92 & +0.09 & 0.67 & 0.75 & +0.08 \\
        \bottomrule
    \end{tabular}
    }
    \caption{Field-level accuracy on CORDv2. 
    \textbf{Separate}: separate extraction, 
    \textbf{Joint}: joint extraction.}
    \label{tab:cordv2_comparison_zeroshot}
\end{table}

\subsection{Evaluation Method}
We measure extraction accuracy\footnote{We manually verify each extraction to accommodate variations in the generated output format. Specifically, a human evaluator compares the model’s predicted string with the ground-truth label, accounting for possible formatting differences (e.g., currency symbols, whitespace). A prediction is considered correct if it matches the intended value (e.g., “3.50” vs. “\$3.50”).} at two levels:
\paragraph{Document-Level Accuracy} For a given document, all queried fields (e.g., Subtotal, Tax, Total) must be correctly extracted to count as a correct document. This strict measure highlights the difficulty of extracting multiple fields correctly at once.

\paragraph{Field-Level Accuracy} Each target field is evaluated independently. This allows us to analyze precisely which fields improve the most (or least) under joint extraction\footnote{We report field-level results only for CORDv2 and SROIE Task~3, as they have consistent field labels amenable 
to grouping and analysis. FUNSD, by contrast, contains noisier or less standardized field names, 
making direct comparison difficult. 
}.

\subsection{Results and Discussion}
\subsubsection{Document-Level Accuracy}
Table~\ref{tab:doc_level_accuracy} presents the document-level accuracy for each model and dataset.
Overall, \textit{joint extraction} achieves higher accuracy than \textit{separate extraction}, 
though the magnitude of the improvements varies by dataset and model. 
The most pronounced gains are observed on CORDv2, where the majority of fields are numeric 
and thus prone to confusion when queried separately, and on FUNSD, which contains rich textual 
information that benefits from the model seeing multiple items in context.
Table~\ref{tab:Joint_examples} shows sample outputs from Llama-3.2-11B-Vision-Instruct 
in the \textit{joint extraction} setting, where the model provides explanatory text and numerical breakdowns 
for each queried field.
In contrast, SROIE Task~3 has relatively fewer and more easily distinguishable fields 
(e.g., \textit{Company Name}, \textit{Address}, \textit{Date}, \textit{Total Amount}), 
leading to smaller performance differences.

\begin{table}[t]
    \centering
    \resizebox{0.48\textwidth}{!}{
    \begin{tabular}{lcccccc}
        \toprule
        \multirow{2}{*}{\textbf{Item}} & \multicolumn{3}{c}{\textbf{Llama-3.2-11B-Vision-Instruct}} & \multicolumn{3}{c}{\textbf{Qwen2-VL-2B-Instruct}} \\
        \cmidrule(lr){2-4} \cmidrule(lr){5-7}
        & \textbf{Separate} & \textbf{Joint} & \textbf{$\Delta$} & \textbf{Separate} & \textbf{Joint} & \textbf{$\Delta$} \\
        \midrule
        Company name  & 0.92 & 0.92 & 0.00  & 1.00 & 1.00 & 0.00 \\
        Address       & 0.98 & 0.95 & -0.03 & 1.00 & 1.00 & 0.00 \\
        Date          & 1.00 & 1.00 & 0.00  & 0.99 & 0.99 & 0.00 \\
        Total Amount  & 0.90 & 0.91 & +0.01  & 0.95 & 0.96 & +0.01 \\
        \bottomrule
    \end{tabular}
    }
    \caption{Field-level accuracy on SROIE Task 3. 
    \textbf{Separate}: separate extraction, 
    \textbf{Joint}: joint extraction.}
    \label{tab:sroie_comparison}
\end{table}

\subsubsection{Field-Level Accuracy}
\label{sec:item-level}

\paragraph{Highly Similar Numeric Fields}
Table~\ref{tab:cordv2_comparison_zeroshot} reports the item-level accuracy on CORDv2, where most target fields are numeric (e.g., \textit{Subtotal}, \textit{Tax}, \textit{Total}). 
\textit{Joint extraction} consistently outperforms \textit{separate extraction}, with particularly notable gains for fields such as \textit{Tax} and \textit{Cash}. 
These fields often appear in tandem on receipts and can be confused with one another or with other monetary values. 
We conjecture that querying them together enables the model to exploit additional cues among these similar numeric entries, helping to reduce confusion\footnote{Table~\ref{tab:json_comparison} shows examples of differences between outputs with separate and joint extractions.}.

\paragraph{Distinct Fields}
Table~\ref{tab:sroie_comparison} shows the item-level accuracy for SROIE, whose fields (\textit{Company Name}, \textit{Address}, \textit{Date}, \textit{Total Amount}) are mostly distinct in both format and surface form.
Consequently, the improvement from \textit{joint extraction} is comparatively smaller. 
Nevertheless, we still observe a modest gain relative to \textit{separate extraction}, suggesting that even loosely related fields may benefit from shared document context.

\begin{table*}[t]
\centering
\small
\begin{tabular}{l|p{13cm}}
\toprule
\multicolumn{2}{c}{\textbf{Sample 1}} \\
\midrule
\textbf{Ground Truth}
& \texttt{\{"subtotal\_price": "1,346,000", "service\_price": "100,950", "tax\_price": "144,695", "total\_price": "1,591,600"\}} \\

\textbf{Separate}
& \texttt{\{"subtotal\_price": "1,346,000.", "service\_price": "100,950.", \textcolor{red}{"tax\_price": "100,950."}, "total\_price": "1,591,600."\}} \\

\textbf{Joint}
& \texttt{\{"subtotal\_price": 1,346,000, "service\_price": 100,950, "tax\_price": 144,695, "total\_price": 1,591,600\}} \\
\midrule
\multicolumn{2}{c}{\textbf{Sample 2}} \\
\midrule
\textbf{Ground Truth}
& \texttt{\{"subtotal\_price": "43.636", "tax\_price": "4.364", "total\_price": "48.000", "cashprice": "50.000", "changeprice": "2.000"\}} \\

\textbf{Separate}
& \texttt{\{"subtotal\_price": "43.636.", "tax\_price": "4.364.", 
\textcolor{red}{"total\_price": "43.636."}, 
\textcolor{red}{"cashprice": "43.636."},
"changeprice": "2.00."\}} \\

\textbf{Joint}
& \texttt{\{"subtotal\_price": 43.636, "tax\_price": 4.364, "total\_price": 48.000, "cashprice": 50.000, "changeprice": 2.000\}} \\
\bottomrule
\end{tabular}
\caption{Model's output comparing \textit{Separate} and \textit{Joint} extraction with the ground truth. 
Fields highlighted in \textcolor{red}{red} differ from the ground truth, illustrating numeric confusion 
that is mitigated under separate extraction.}
\label{tab:json_comparison}
\end{table*}

\begin{table*}[t]
\centering
\resizebox{\textwidth}{!}{
\begin{tabular}{cccccccccc}
\toprule
\multirow{2}{*}{\textbf{Target Field}} & \multirow{2}{*}{\textbf{Joint-Inference Fields}} & \multirow{2}{*}{$R^2$} & \multirow{2}{*}{\textbf{n}} & \multicolumn{3}{c}{\textbf{Llama-3.2-11B-Vision-Instruct}} & \multicolumn{3}{c}{\textbf{Qwen2-VL-2B-Instruct}} \\ \cline{5-10}
 &  &  &  & \textbf{Separate} & \textbf{Joint} & \textbf{$\Delta$} & \textbf{Separate} & \textbf{Joint} & \textbf{$\Delta$} \\ \hline
\multirow{2}{*}{Tax} &Subtotal, Total & 0.99 & \multirow{2}{*}{80} & 0.60 & 0.80 & \textbf{+0.20} & 0.69 & 0.80 & \textbf{+0.11} \\ \cline{2-3} \cline{5-10} 
 & Change, Menu Types & 0.05 &  & 0.60 & 0.59 & -0.01 & 0.69 & 0.68 & -0.01 \\ \hline
\multirow{2}{*}{Cash} & Total, Change & 0.95 & \multirow{2}{*}{100} & 0.72 & 0.81 & \textbf{+0.09} & 0.71 & 0.86 & \textbf{+0.15} \\ \cline{2-3} \cline{5-10} 
 & Change, Menu Types & 0.02 &  & 0.72 & 0.69 & -0.03 & 0.71 & 0.60 & -0.11 \\ \hline
\multirow{2}{*}{Change} & Total, Cash & 0.98 & \multirow{2}{*}{100} & 0.84 & 0.90 & \textbf{+0.06} & 0.83 & 0.92 & \textbf{+0.09} \\ \cline{2-3} \cline{5-10} 
 &Subtotal, Tax & 0.04 &  & 0.84 & 0.82 & -0.02 & 0.83 & 0.84 & +0.01 \\
 \bottomrule
\end{tabular}
}
\caption{
Dependency analysis of numeric fields in CORDv2. 
\(\textbf{n}\) denotes the number of triplets. 
}
\label{tab:interdependency_results}
\end{table*}

\section{Analysis}
In this section, we delve deeper into the conditions under which \textit{joint extraction} outperforms \textit{separate extraction}. 
First, we quantitatively measure the dependencies among numeric fields (Section~\ref{subsec:dependencies}), 
showing how strong field interdependence contributes to performance gains in joint extraction. 
Next, we examine how performance changes when the number of queried fields increases (Section~\ref{subsec:number_of_items}).

\subsection{Impact of Field Dependencies}
\label{subsec:dependencies}
\subsubsection{Definition of Dependencies}
We define a dependency metric using the training set of CORDv2 dataset, focusing on triplets of fields \((x, y, z)\). 
Here, \(x\) is the primary field of interest (e.g., \textit{Tax}), and \(y\) and \(z\) are two related fields 
(e.g., \textit{Subtotal} and \textit{Total}). We fit the following multiple linear regression model on the training set:
\[
x = c_1 y + c_2 z + b,
\]
where \(c_1, c_2\), and \(b\) are trainable coefficients. The \textit{coefficient of determination} \((R^2)\) 
indicates how well \(x\) can be explained by \(y\) and \(z\); a higher \(R^2\) implies stronger dependency.
We define:
\begin{itemize}
    \item High-\(R^2\) (\(R^2 \geq 0.9\)): \(x\) is highly predictable from \(y\) and \(z\).
    \item Low-\(R^2\) (\(R^2 \leq 0.1\)): \(x\) has little to no linear relationship with \(y\) and \(z\).
\end{itemize}

Our hypothesis is that when fields share high interdependence (High-\(R^2\)), 
\textit{joint extraction} should yield a significant advantage over \textit{separate extraction} 
by leveraging contextual cues among the fields. Conversely, if fields are largely independent (Low-\(R^2\)), 
we expect minimal gains from querying them jointly.

\subsubsection{Results}
Table~\ref{tab:interdependency_results} reports the results for 
Llama-3.2-11B-Vision-Instruct and Qwen2-VL-2B-Instruct. 
Under High-\(R^2\) conditions, \textit{joint extraction} substantially outperforms 
\textit{separate extraction}, indicating that strong numeric dependencies facilitate more effective disambiguation when multiple fields are processed together. 
In contrast, for Low-\(R^2\) fields, both methods perform comparably, suggesting that \textit{joint extraction} provides limited benefit when fields lack strong correlation.

These findings align with our earlier observations: 
multi-field reasoning is most beneficial when target fields exhibit considerable interdependence. 
By quantifying these dependencies, we can pinpoint the scenarios in which joint extraction delivers the greatest impact, offering a data-driven guideline for practical deployments.

\subsection{Effect of Increasing the Number of Target Fields}
\label{subsec:number_of_items}
Many real-world applications require extracting multiple fields simultaneously. 
To examine how \textit{joint extraction} scales, we vary the number of fields in a single prompt (from two to six) using the CORDv2 development and test sets. 
We group the receipts according to how many labeled fields each contains, then measure whether \textit{all} specified fields are correctly extracted.

\begin{figure}[t]
\centering
\includegraphics[width=0.48\textwidth]{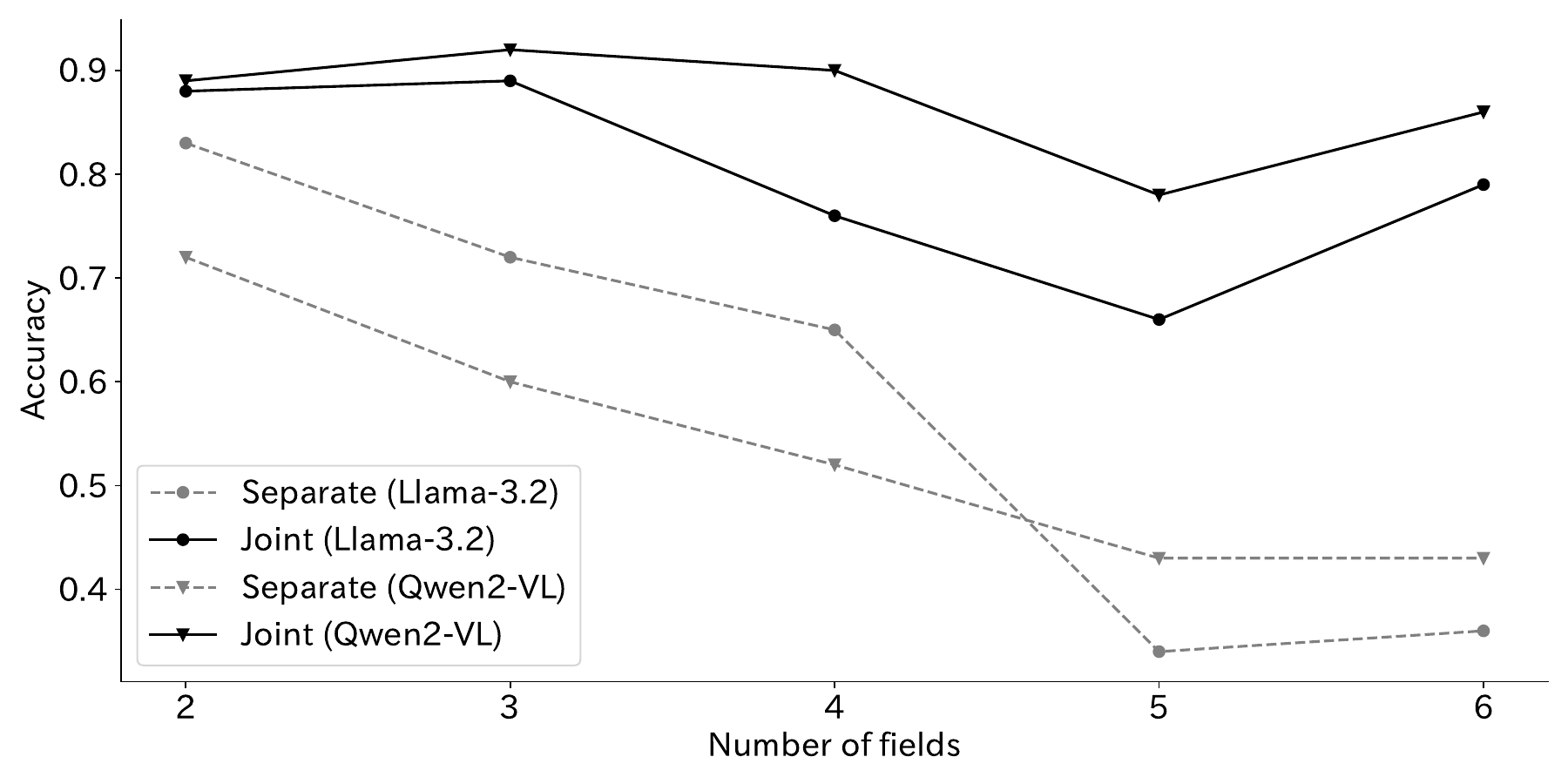}
\caption{
Document-level accuracy on CORDv2 as the number of queried fields increases from two to six. \textit{Joint extraction} remains comparatively robust, while \textit{separate extraction} shows a more pronounced decline in accuracy for larger field sets.}
\label{fig:items_vs_accuracy}
\end{figure}

Figure~\ref{fig:items_vs_accuracy} presents the document-level accuracy as the number of fields increases. 
Across all evaluated conditions, \textit{joint extraction} maintains higher accuracy 
than \textit{separate extraction}, even as the query size grows. 
In contrast, \textit{separate extraction} accuracy declines more rapidly for larger numbers of fields, likely due to errors compounding with each additional independent query. 
\textit{Joint extraction} remains comparatively robust, indicating that a single prompt encompassing multiple fields helps the model retain context and reduce confusion.


\section{Conclusion}
We investigated the performance of \textit{joint} versus \textit{separate} extraction strategies for document information extraction using prompt-based VQA. Through extensive experiments on various datasets and models, we demonstrated that extracting multiple fields together often boosts accuracy, particularly when the fields are numerically or contextually interdependent. These findings underscore the practical advantages of employing a single multi-field prompt, especially for real-world scenarios involving closely related fields or numerous target items.

\section{Limitations}
While \textit{joint extraction} achieves higher accuracy by leveraging inter-field dependencies, the question of inference time remains. 
As the number of requested fields grows, a single prompt may become longer, potentially increasing computational overhead. Conversely, \textit{separate extraction} requires multiple prompts—one per field—introducing repeated overhead that can also inflate inference costs. 
A more extensive study of latency, memory usage, and scalability to larger or more complex extractions remains outside the scope of this work.

\bibliography{acl_latex, anthology}

\newpage
\appendix
\section{Few-Shot Setting}
\label{appendix:few-shot}
In the main paper, we focus on zero-shot experiments to isolate model capacity and avoid biases from few-shot prompt engineering. 
For reference, Table~\ref{tab:cordv2_comparison} shows results under a 4-shot setting, 
where a small number of labeled examples are included with each query prompt. 
Although few-shot examples can enhance performance, they also introduce variability in how examples are selected and presented. 
Our overall conclusions remain consistent: \textit{joint extraction} provides tangible benefits over 
\textit{separate extraction} when the fields share numerical or contextual cues.

\begin{table}[h!]
    \centering
    \resizebox{0.48\textwidth}{!}{%
    \begin{tabular}{lcccccc}
        \toprule
        & \multicolumn{6}{c}{\textbf{4-shot}} \\
        \cmidrule(lr){2-7}
        \textbf{Item}
        & \multicolumn{3}{c}{\textbf{Llama-3.2-11B-Vision-Instruct}}
        & \multicolumn{3}{c}{\textbf{Qwen2-VL-2B-Instruct}} \\
        \cmidrule(lr){2-4} \cmidrule(lr){5-7}
        & \textbf{Separate} & \textbf{Joint} & \textbf{$\Delta$}
        & \textbf{Separate} & \textbf{Joint} & \textbf{$\Delta$} \\
        \midrule
        Subtotal    & 0.68 & 0.67 & -0.01 & 0.90 & 0.95 & +0.05 \\
        Total       & 0.69 & 0.69 &  0.00 & 0.94 & 0.98 & +0.04 \\
        Tax         & 0.16 & 0.29 & +0.13 & 0.84 & 0.97 & +0.13 \\
        CreditCard  & 0.73 & 0.80 & +0.07 & 0.80 & 0.80 & 0.00 \\
        Quantity    & 0.52 & 0.52 &  0.00 & 1.00 & 1.00 & 0.00 \\
        Cash        & 0.82 & 0.83 & +0.01 & 0.78 & 0.97 & +0.19 \\
        Change      & 0.71 & 0.77 & +0.06 & 0.92 & 0.96 & +0.04 \\
        Service     & 0.25 & 0.25 &  0.00 & 0.92 & 0.92 & 0.00 \\
        \midrule
        Overall     & 0.45 & 0.64 & +0.19 & 0.69 & 0.89 & +0.20 \\
        \bottomrule
    \end{tabular}}
    \caption{
    Field-level accuracy on the CORDv2 dataset under the 4-shot setting. 
    \textbf{Separate}: separate extraction, 
    \textbf{Joint}: joint extraction.
    }
    \label{tab:cordv2_comparison}
\end{table}


\end{document}